\documentclass[10pt,twocolumn,letterpaper]{article}

\usepackage{iccv}
\usepackage{times}
\usepackage{epsfig}
\usepackage{graphicx}
\usepackage{amsmath}
\usepackage{amssymb}
\usepackage{multirow}
\usepackage{subcaption}

\usepackage{xcolor}
\definecolor{river}{RGB}{0, 0, 200}
\definecolor{lake}{RGB}{0, 150, 250}
\definecolor{pond}{RGB}{0, 150, 200}
\definecolor{dred}{RGB}{200, 0, 0}
\definecolor{dgreen}{RGB}{44, 160, 44}
\renewcommand\fbox{\fcolorbox{dred}{white}}
\newcommand\crule[3][black]{{\setlength{\fboxsep}{0pt}\fbox{\textcolor{#1}{\rule{#2}{#3}}}}}
\newcommand\blackfbox{\fcolorbox{black}{white}}
\newcommand\blackcrule[3][black]{{\setlength{\fboxsep}{0pt}\blackfbox{\textcolor{#1}{\rule{#2}{#3}}}}}

\usepackage[breaklinks=true,bookmarks=false]{hyperref}

\iccvfinalcopy 

\def\iccvPaperID{1646} 

\ificcvfinal\fi

\begin{document}

\title{Seeing Beyond the Patch: Scale-Adaptive Semantic Segmentation of High-resolution Remote Sensing Imagery based on Reinforcement Learning}

\author{Yinhe Liu, Sunan Shi, Junjue Wang, Yanfei Zhong\thanks{corresponding author.}
\\
Wuhan University,
Wuhan, China\\
{\tt\small \{liuyinhe, shisunan, kingdrone, zhongyanfei\}@whu.edu.cn}
}


\maketitle
\ificcvfinal\thispagestyle{empty}\fi

\begin{abstract}
   In remote sensing imagery analysis, patch-based methods have limitations in capturing information beyond the sliding window. This shortcoming poses a significant challenge in processing complex and variable geo-objects, which results in semantic inconsistency in segmentation results.
   To address this challenge, we propose a dynamic scale perception framework, named GeoAgent, which adaptively captures appropriate scale context information outside the image patch based on the different geo-objects. In GeoAgent, each image patch's states are represented by a global thumbnail and a location mask. The global thumbnail provides context beyond the patch, and the location mask guides the perceived spatial relationships.
   The scale-selection actions are performed through a Scale Control Agent (SCA). A feature indexing module is proposed to enhance the ability of the agent to distinguish the current image patch's location. The action switches the patch scale and context branch of a dual-branch segmentation network that extracts and fuses the features of multi-scale patches. The GeoAgent adjusts the network parameters to perform the appropriate scale-selection action based on the reward received for the selected scale.
   The experimental results, using two publicly available datasets and our newly constructed dataset WUSU, demonstrate that GeoAgent outperforms previous segmentation methods, particularly for large-scale mapping applications.
\end{abstract}

\vspace{-0.7cm}
\section{Introduction}
\vspace{-0.2cm}
The acquisition of high spatial resolution (HSR) remote sensing images has become possible with the advancement of sensors \cite{6198307}. These images enable the extraction of geospatial objects, such as buildings, roads, and aircraft, through image interpretation \cite{xia2018dota, waqas2019isaid, Li_2019_ICCV, Zhou_2018_CVPR_Workshops}. Furthermore, HSR remote sensing images facilitate the determination of more semantic land use/ land cover (LULC) categories, including industrial land, cropland, and residential land \cite{demir2018deepglobe}. Such LULC mapping is critical in urban management, planning, and ecological conservation \cite{bechtel2015mapping, huang2018urban, murray2022high}.

The task of LULC mapping requires the semantic categorization of each pixel, also known as semantic segmentation. Recently, Fully Convolutional Networks (FCNs) \cite{long2015fully} have achieved significant progress for natural images in computer vision \cite{wang2021exploring, xie2021segformer, huang2019ccnet, unet, pspnet, deeplabv3p}. Deep learning techniques have also shown success in the automatic interpretation of HSR remote sensing images \cite{9174822, ding2021object, Zheng_2020_CVPR, Li_2021_CVPR, Mou_2019_CVPR}.

\begin{figure}[t]
  \centering
  \includegraphics[width=1.0\linewidth]{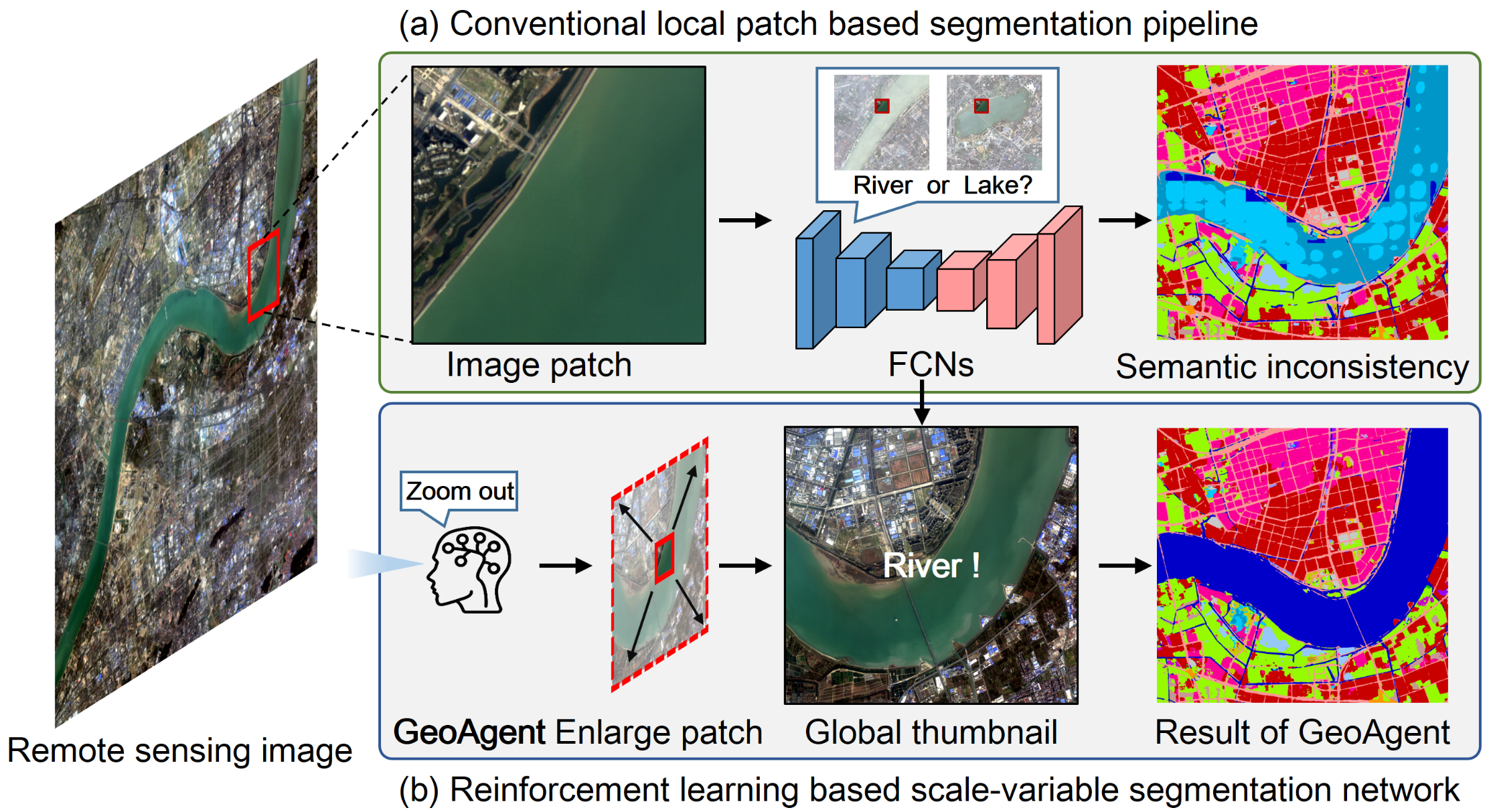}
  \vspace{-0.5cm}
  \caption{Illustration of the scale problem for HSR remote sensing image LULC segmentation. The sliding window \crule[white]{0.2cm}{0.2cm} is enlarged for a better view. Legend: \blackcrule[river]{0.2cm}{0.2cm} river, \blackcrule[lake]{0.2cm}{0.2cm} lake, \blackcrule[pond]{0.2cm}{0.2cm} pond.}
  \label{fig:problem}
  \vspace{-0.5cm}
\end{figure}
In contrast to conventional ground-based images, HSR remote sensing images have a large length and width that can encompass thousands of objects in a single image. However, distinguishing LULC (Land Use and Land Cover) geo-objects that vary greatly at different scales using a fixed, finite size image patch is challenging. Although Fully Convolutional Networks (FCNs) can theoretically receive input of arbitrary size, when processing HSR remote sensing images, a sliding window (SW) of a specific size is often used to generate image patches \cite{wang2022cross,van2018spacenet,diakogiannis2020resunet}.
In small patches, different LULC classes may consist of the same objects with similar patterns, making them almost indistinguishable \cite{zhong2020open, xia2017aid}. For instance, the same water may appear in rivers, lakes, and ponds, and the same dense housing may belong to rural or urban residential land. Moreover, some cultivated land and meadows may have similar soils \cite{tong2020land,tong2022enabling}. Regardless of the multi-scale module's design \cite{deeplabv3p, FPN,liu2015parsenet, pspnet,dosovitskiy2020image, wang2018non, liu2021swin}, the model's receptive field cannot surpass the image patch's size \cite{araujo2019computing}.
Consequently, the commonly used Deeplabv3+ \cite{deeplabv3p} model's mapping results exhibit an apparent grid effect due to the gap between the sliding window's limited size and the river's enormous scale, as illustrated in Figure \ref{fig:problem}. 
\par
The accurate segmentation of land use and land cover (LULC) in high spatial resolution (HSR) remote sensing images poses a challenge due to the significant scale variance and limited receptive field of the network by patch size. Various models aim to address this issue by tackling the multiscale problem from the perspective of input data \cite{chen2019collaborative, WiCoNet, cascadepsp}. While incorporating contextual information can effectively resolve the problem of identifying rivers and water bodies within limited image patches, different LULC geo-objects necessitate different scales of image patches. For instance, local image patches or slightly larger scales may suffice for identifying a pit pond, whereas lakes and continuous rivers may require larger scales. An inappropriate scale of receptive field may have an adverse impact on the segmentation.
\par
The supervised learning methods have difficulty in adaptively determining the appropriate patch scales, as it requires manual labeling of images that correspond to the appropriate context scale. In contrast, reinforcement learning (RL) algorithms operate through interactions between the \textit{agent} and the \textit{environment}, receiving rewards for \textit{actions} and using that information to update model parameters, without requiring labeled data \cite{sutton2018reinforcement}. RL techniques have achieved success in various areas, including board games, video games, and mathematics \cite{mnih2013playing, silver2018general, fawzi2022discovering}. Thus, applying deep RL techniques to LULC segmentation networks may allow for the adaptive adjustment of the receptive field scale based on image features, dynamically controlling the data flow of HSR remote sensing images.
Therefore, we propose a scale-adaptive segmentation network, called GeoAgent to dynamically select appropriate patch scale based on different geo-obejcts. A HSR remote sensing segmentation dataset WUSU is constructed for experiment. The main contributions include:

(1) A RL-based HSR imagery segmentation method is proposed. In GeoAgent, the HSR remote sensing images are represented by different \textit{states}. GeoAgent observes different \textit{states} and performs a series of scale-selection \textit{actions} to control the area of each patch, obtaining the final global mapping \textit{reward} based on the selected scales. 

(2) A Scale Control Agent (SCA) is proposed to dynamically controls the scale of the GeoAgent. The SCA contains two parts: an 'actor' network that learns a policy to map the \textit{states} to a probability distribution of \textit{actions} and a 'critic' network that evaluates the value of each \textit{action} using an advantage function. A feature indexing module is proposed to enhance the \textit{agent}'s ability to distinguish the location of the current image patch. 

(3) GeoAgent use a dual-branch segmentation network for multi-scale image patches segmentation. The input scale and context branch are switched under the control of the \textit{actions} selected by SCA. The network processes the multi-scale context patches and local scale patches separately, and the multi-scale features are aggregated based on the geo-location information to generate the final multi-scale segmentation result.

\vspace{-0.3cm}
\section{Related Works}
\vspace{-0.2cm}
\subsection{Multi-Scale Segmentation Networks}
\vspace{-0.1cm}

There have been several techniques to solve multi-scale problems, ranging from multi-scale feature ensemble (e.g. Deeplab \cite{deeplabv3p}, FPN \cite{FPN}) to context information preservation (e.g. ParseNet \cite{liu2015parsenet}, PSPNet \cite{pspnet}).
Through the self-attention mechanism, it is able to capture long-distance relationships in images \cite{dosovitskiy2020image,wang2018non,liu2021swin}.
But no matter how the multi-scale module is designed, the receptive field of the model cannot break through the size of the sliding window. 
\begin{figure*}[!t]
  \centering
  \includegraphics[width=1.0\linewidth]{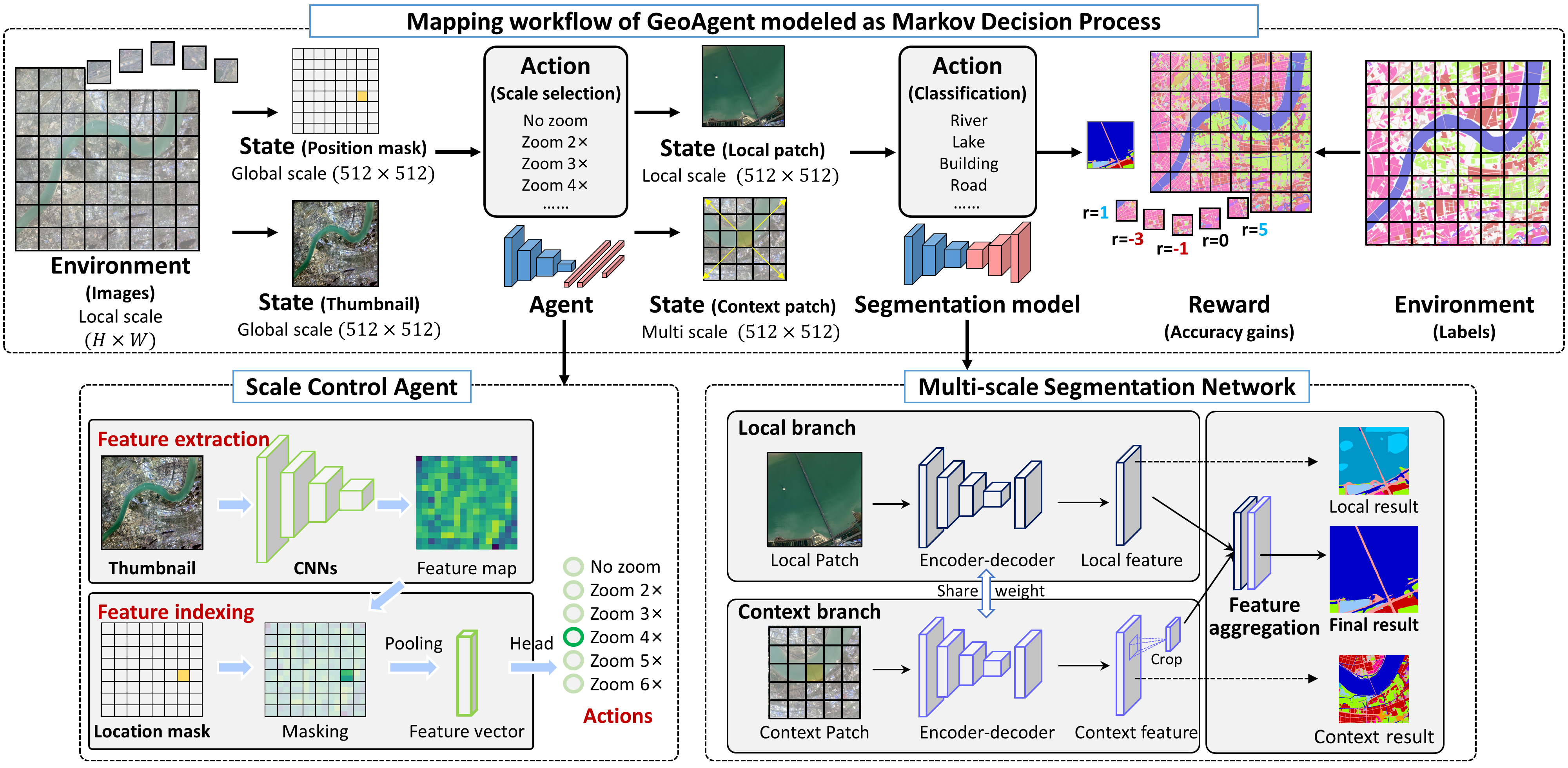}
  \vspace{-0.5cm}
  \caption{The overall flowchart of the proposed reinforcement learning based scale-adaptive segmentation network GeoAgent.}
  \label{fig:overall}
  \vspace{-0.5cm}
\end{figure*}
\par
Therefore some models try to solve the multi-scale problem from the perspective of input data.
\cite{recasens2018learning} proposed a saliency-based distortion layer for convolutional neural networks to improve the spatial sampling of input data with the condition of the given task.
Some methods fetch larger scale image outside the patch to provide spatial context.
\cite{chen2019collaborative} proposed the GLNet, which consists of a global branch and a local branch, which take the downsampled global image and the full-resolution cropped image as input, respectively, effectively preserving the detail information and the global context information.
\cite{WiCoNet} propose a wide-context network (WiCoNet), a context transformer is introduced to extract the downsampled global context image. 
On the other hand, some approaches address the problem by leveraging multi-scale processing.
\cite{cascadepsp} proposed CascadePSP, which first performs segmentation on the downsampled coarse resolution images, and then uses the original resolution images to refine the segmentation results.
\cite{MagNet} proposed multi-scale framework MagNet that uses multiple processing stages to resolve local ambiguity by examining the image at various magnification levels and propagating coarse-to-fine information.
However, some of the above methods do not break the limits of image patches \cite{recasens2018learning}, or introduced a fixed scale context image\cite{chen2019collaborative,WiCoNet} and complex processes \cite{MagNet,cascadepsp} that do not allow the model adaptively choose appropriate scale to zoom the sliding window based on different objects.
\vspace{-0.1cm}
\subsection{DRL-based Dynamic Networks}
\vspace{-0.1cm}
Deep Reinforcement Learning (DRL) is able to autonomously make decisions about the following \textit{action} by observing the \textit{state} of \text{environment}, which in turn affects the \text{environment}, and updating the model parameters by receiving the \text{environment}'s \textit{reward} for the \textit{action}. By introducing the DRL approach, CNNs can be more dynamic and able to automatically change the network structure based on the input or influence the input to the network.
\cite{wu2018blockdrop} uses reinforcement learning to train a policy network for double \textit{reward}. During training and testing time, the \textit{agent} uses the input image as a \textit{state} to try to utilize the least number of blocks while maintaining recognition accuracy.
\cite{uzkent2020learning} proposed a Patch-Drop algorithm to selectively use high-resolution satellite data when necessary. A policy network is utilized to take low-resolution remote sensing images as \textit{state} to decide whether to sample high-resolution remote sensing images as additional inputs.
\cite{ayush2021efficient} proposed a reinforcement learning setup to acquire high-resolution satellite images conditionally. A cost-aware \textit{reward} function was designed to reduce satellite image acquisition costs while maintaining accuracy.
\cite{RAZN} proposed RAZN to achieve accurate and fast prediction of breast cancer segmentation in whole-slide images by learning a policy network to decide whether zooming is required in a given region of interest, motivated by the zoom-in operation of a pathologist using a digital microscope.
To the authors' knowledge, there has yet to be work using RL to solve the problem of limited patch size in remote sensing imagery segmentation task, so the scale-variable network GeoAgent is proposed.

\vspace{-0.1cm}
\section{Methodology}
\vspace{-0.1cm}
\subsection{Overview}
\vspace{-0.1cm}
The GeoAgent for HSR remote sensing image segmentation comprises two sub-networks: the Scale Control Agent (SCA) $\pi(a|s)$ and the multi-scale segmentation network $\pi(Y|X)$. The entire HSR image is considered as the segmentation target, and the segmentation process is modeled as a Markov Decision Making (MDP) process. The SCA selects an appropriate scale based on the \textit{states} $s$ of global thumbnail $X^{\downarrow}_{thumb}$ and the position mask $M_{pos}$, and determines the \textit{action} $a$ which control the input scale of the segmentation network. The dual-branch segmentation network $\pi(Y|X)$ takes the local image patches $X_{loc}$ and the multi-scale context image patch $X^{\downarrow }_{context}$ to assign categories to each pixel of the current patch. The sequence of \textit{states}, \textit{actions}, and \textit{rewards} form a complete \textit{trajectory} $\tau $, representing the entire mapping process of image $X_{i} $. The large scale segmentation results $\hat{Y}_{i} $ are obtained and a final mapping \textit{reward} will be given. The objective of GeoAgent is to learn an \textit{agent} $\pi(a|s)$ and a segmentation network $\pi(Y|X)$ that maximize the expected expected cumulative \textit{reward} of $J(\tau )$.
\vspace{-0.1cm}
\subsection{Scale Control Agent Module}
\vspace{-0.1cm}
\subsubsection{Patch Scale Selection}
\vspace{-0.1cm}
The High Spatial Resolution (HSR) remote sensing image segmentation dataset, denoted as ${\mathcal D}=\left\{\left(X_{i} ,Y_{i} \right)\right\}_{i=1}^{N} $, comprising $N$ pairs of HSR remote sensing images $X{i}$ and the corresponding training labels $Y_{i}$, serves as the RL \text{environment} in this study. Each \textit{trajectory} $\tau$ begins with the \text{environment}'s initialization, achieved by randomly selecting an image $X_{i}$ of size $H\times W$ at its original resolution, which is subsequently divided into $T$ fixed-size image patches. Each image patch's segmentation defines a timestep $t$ in the \textit{trajectory}.
\par
At each timestep, the \textit{state} $s_{t}$ is represented by the global thumbnail $X^{\downarrow}_{thumb}$ and the position mask $M_{pos}$ of the current patch. The thumbnail $X^{\downarrow}_{thumb}$ is a downsampled HSR image of specific size $h\times w$ (e.g., $512\times 512$), containing the global-scale context under coarse resolution. The position mask $M_{pos}$ is a binary mask of size $h\times w$, indicating the current patch's position, with the value of the current extent set to 1 and the rest of the pixels set to 0.
\par
As shown in Fig.\ref{fig:overall}, the Scale Control Agent (SCA) $\pi(a|s)$ is comprised of a policy network, namely, the 'actor', that determines the appropriate \textit{action} $a_{t}$ for each timestep (image patch) $t$. The \textit{action} space $a_{t} \in 1,2,\ldots ,N$ represents the scale of the context. In addition to the local image patches $X_{loc}$ of size $h\times w$ with the original resolution, the context image patch $X^{\downarrow}_{context}$ of same size $h\times w$ with coarse resolution downsampled by a factor of $a{t}$ is acquired. The segmentation network $\pi(Y|X)$ assigns categories to each pixel of the current patch based on $X_{loc}$ and $X^{\downarrow}_{context}$. The \textit{action} $a_t$ receives an immediate \textit{reward} $r{t}$ based on the segmentation result and the \textit{environment} $Y_{i}$. Following the \textit{action}, the \textit{environment} transitions to the next \textit{state}, where the sliding window moves to the next image patch. This sequence of \textit{actions} is repeated until the entire image is observed, and the final mapping results $\hat{Y}_{i}$ are obtained.
\par
Since the difference between the \textit{states} of the different positions lies only in position mask $M_{pos} $, this makes discriminability of the different \textit{states} very small. When the \textit{states} are directly used as input, it is difficult for SCA to recognize the weak differences and thus select similar \textit{actions}. To improve the learning capability of the SCA, a feature indexing module is proposed to handle the subtle differences between \textit{states}. As shown in Fig.\ref{fig:overall}, the SCA is not directly trained to learn the mapping from similar \textit{states} to different \textit{actions}. First, the thumbnail $X^{\downarrow }_{thumb} $ is fed to a CNN-based backbone for whole-image feature extraction. Then, the features relevant to the current patch are indexed by the position mask $M_{pos} $, and the irrelevant features outside the patch are masked. The features of the current patch are extracted through a $3\times 3$ window and transformed into feature vectors by global average pooling. Finally, an actor head with multiple fully connected layers outputs the \textit{action} probabilities.
\par
\vspace{-0.3cm}
\subsubsection{Segmentation Rewards and Values of Actions}
\vspace{-0.1cm}
\par
After formulating the MDP of GeoAgent, the \textit{agent}'s parameters must be optimized using DRL methods to maximize the cumulative \textit{reward} $J(\tau)$. 
The goal of the SCA is to select the appropriate scale to obtain contextual information outside the patch and improve mapping accuracy. To penalize inappropriate \textit{actions}, the immediate \textit{reward} for each patch is defined as the accuracy gain of the final result based on the \textit{action} $a_{t} $ chosen by the \textit{agent} compared to the result obtained only by the local scale, i.e., $a_{t} =1$: 
\begin{equation}
  \setlength{\abovedisplayskip}{3pt}
  \setlength{\belowdisplayskip}{3pt}{\rm {\mathcal R}}_{patch} (a_{t} ,y)=Score(y,\hat{y}(a_{t} ))-Score(y,\hat{y}(1)) 
\end{equation} 
where $Score(y,\hat{y})=mIoU(y,\hat{y})+mF_{1}(y,\hat{y})$, and $mIoU$ and $mF_{1} $ are mean Intersection-Over-Union and mean F${}_{1}$ score, respectively. The \textit{reward} value based only on the local image is 0, and the value is positive if the \textit{action} leads to an accuracy improvement and negative if it results in a decline in accuracy. However, rewarding only the current single patch may cause the \textit{agent} to overlook the potential long-term benefits of whole image mapping. Therefore, after completing the episode, the accuracy of the predicted results for the entire image is evaluated, and the final global mapping \textit{reward} is given: 
\begin{equation}
  \setlength{\abovedisplayskip}{3pt}
  \setlength{\belowdisplayskip}{3pt}{\rm {\mathcal R}}_{map} =T*(Score(Y_{i} ,\hat{Y}_{i} )-Score(Y_{i} ,\hat{Y}_{i}^{1} )) 
\end{equation} 
where $T$ is the number of generated patches for normalization, and $\hat{Y}_{i}^{1} $ is the mapping result of the whole image $X_{i} $ based only on the local scale.
\par
To evaluate the expected accumulative future \textit{reward} of a \textit{state} or \textit{state}-\textit{action} pair, \textit{value} function $V^{\pi } (s)$ and $Q^{\pi } (s,a)$ are introduced: $V^{\pi } (s)$ denotes the cumulative \textit{reward} from using policy $\pi $ from \textit{state} $s$: 
\begin{equation} \label{2)} 
V^{\pi } (s)={\rm {\rm E}}_{s_{0} =s,\tau {\rm \sim }\pi } \left[\sum _{t=0}^{T} \gamma ^{t} r_{t} \right] 
\end{equation} 
and $Q^{\pi } (s,a)$ denotes the cumulative \textit{reward} brought by performing \textit{action} $a$ from \textit{state} $s$ based on policy $\pi $: 
\begin{equation} \label{3)} 
Q^{\pi } (s,a)={\rm {\rm E}}_{s_{0} =s,a_{0} =a,\tau {\rm \sim }\pi } \left[\sum _{t=0}^{T} \gamma ^{t} r_{t} \right] 
\end{equation} 
\par
To estimate the \textit{value} of each \textit{action}, in addition to the actor network to learn the mapping from current \textit{states} to \textit{actions}, SCA also contains a critic network that shares weights with the actor network to learn the \textit{value} function for evaluating the current \textit{states}and \textit{actions}. The critic network shares a CNN backbone with the actor network to extract the features, but has a separate header to output the estimated values for later optimization of the network parameters.
\vspace{-0.1cm}
\subsection{Scale-Variable Segmentation based on Dual-Branch Network}
\vspace{-0.1cm}
The Dual-Branch segmentation Network, denoted as $\pi(Y|X)$, produces the segmentation \textit{action} given the \textit{state} of a local image patch $X_{loc} $ and a multi-scale context patch $X^{\downarrow }_{context} $. The local patch $X_{loc} $, of size $h\times w$, is sampled from the original resolution image based on the patch position. The context patch $X^{\downarrow }_{context} $ is obtained by expanding the local patch size by a factor of $a_{t} $ centered at the current position and then downsampling it by the same factor to obtain a context patch of size $h\times w$.

As shown in Fig.\ref{fig:overall},
the $\pi(Y|X)$ network employs a two-branch structure sharing a common segmentation network to handle multi-scale input images. The context branch takes downsampled global thumbnails $X^{\downarrow }_{context} $, while the local branch receives local patches $X_{loc} $ of the same size. When $a_{t} =1$, i.e., when the \textit{agent} believes that only local image patches are sufficient, the context branch is deactivated, and the dual-branch segmentation network is reduced to a single branch network. During the segmentation process, all layers in either branch share the same weights. The high-level feature maps $f_{l} $ and $f_{c} $ are then combined to generate the final segmentation mask using a branch aggregation layer.


During the feature fusion phase, the range corresponding to the local image is cropped from the context patch feature map based on its geographic coordinates. The two scales of feature maps are concatenated and fused by a convolutional layer of size $3\times 3$. Finally, the segmentation probability is computed using a $1\times 1$ convolution and a softmax layer. To facilitate training for both branches, auxiliary headers are added to output segmentation results for local patches and context patches based on either $f_{l} $ or $f_{g} $ at a single scale. 
\vspace{-0.1cm}
\subsection{Optimization of GeoAgent}
\vspace{-0.1cm}
The Advantage Actor Critic (A2C) algorithm is employed for stable learning of the \textit{agents} in an adaptive \textit{environment} \cite{A2C}. As illustrated in Section 3.2, the SCA module consists of two learnable components: an "actor" $\theta _{A} $ that learns the parameterized policy and a "critic" $\theta _{C} $ that learns the \textit{value} function to evaluate the \textit{state}-\textit{action} pair. The critic provides reinforcement signals to the actor by learning the advantage function $A^{\pi } (s,a)$ \cite{A2C}: 
\begin{equation}
  \setlength{\abovedisplayskip}{3pt}
  \setlength{\belowdisplayskip}{3pt}A^{\pi } (s,a)=Q^{\pi } (s,a)-V^{\pi } (s) 
\end{equation} 
where $V^{\pi } (s)$ denotes the cumulative \textit{reward} from using policy $\pi $ from \textit{state} $s$, and $Q^{\pi } (s,a)$ denotes the cumulative \textit{reward} brought by performing \textit{action} $a$ from \textit{state} $s$ based on policy $\pi $. 

The target advantage value $\hat{V}_{tar}^{\pi } (s_{t} )$ can be estimated by the temporal difference (TD) algorithm and \textit{trajectory} data (Sutton 1988), and used to train the critic network $\theta _{c} $ by mean squared error (MSE) loss \cite{A2C}: 
\begin{equation}
  \setlength{\abovedisplayskip}{3pt}
  \setlength{\belowdisplayskip}{3pt}L_{value} (\theta _{C} )=\frac{1}{T} \sum _{t=0}^{T} (\hat{V}^{\pi } (s_{t} )-V_{tar}^{\pi } (s_{t} ))^{2}  
\end{equation} 

The actor network predict the advantage \textit{value} $\hat{A}^{\pi } (s_{t} ,a_{t} )$ and optimized by the advance based on the policy loss (Mnih et al. 2016; Williams 1992): 
\begin{equation}
  \setlength{\abovedisplayskip}{3pt}
  \setlength{\belowdisplayskip}{3pt}L_{policy} (\theta _{A} )=\frac{1}{T} \sum _{t=0}^{T} (-\hat{A}^{\pi } (s_{t} ,a_{t} )\log \pi \theta _{A} (a_{t} |s_{t} )) 
\end{equation} 

The parameters of the segmentation network $\pi(Y|X)$ can be updated through supervised learning from the loss function calculated directly from the labels. The commonly used cross-entropy loss function is employed to optimize the segmentation network. In addition to the loss values computed from the final outputs, additional segmentation head outputs in the global and local branches are also used to compute the auxiliary losses. 

When the entire GeoAgent is trained, the segmentation network $\pi(Y|X)$ is first pre-trained by taking random \textit{action} $a_{t} $. After completing the pre-training, the SCA $\pi(a|s)$ is trained to learn to select the appropriate scale dynamically. Finally, joint learning is performed to optimize the segmentation network and the SCA. Since the data reading pipelines of the two training methods differ greatly, $\theta _{scale} $ and $\theta _{cls} $ are updated asynchronously at specific intervals to avoid the instability caused by simultaneous optimization.

\section{Experiments}
\subsection{Experimental Setup}
\vspace{-0.1cm}
\begin{table*}[!htb]
  \centering
  \vspace{-0.2cm}
  \caption{Segmentation accuracies of different models on three datasets.}
  \vspace{-0.2cm}
  \label{table:results}
  \resizebox{1.9\columnwidth}{!}{
      \begin{tabular}{llllllll} 
      \hline
      \multirow{2}{*}{Method} & \multirow{2}{*}{Backbone} & \multicolumn{2}{c}{WUSU}                                                    & \multicolumn{2}{c}{GID}                                                     & \multicolumn{2}{c}{FBP}                                                      \\ 
      \cline{3-8}
                              &                           & IoU(\%)                              & F$_1$(\%)                            & IoU(\%)                              & F$_1$(\%)                            & IoU(\%)                              & F$_1$(\%)                             \\ 
      \hline
      \multicolumn{8}{c}{Local scale patch based methods}                                                                                                                                                                                                                                                                   \\ 
      \hline
      -                       & UNet \cite{unet}                     & 54.10                                & 59.52                                & 58.76                                & 68.93                                & 53.02                                & 59.67                                 \\
      -                       & PSPNet   \cite{pspnet}                 & 60.37                                & 69.29                                & 59.04                                & 69.34                                & 59.69                                & 64.27                                 \\
      -                       & FPN    \cite{FPN}                   & 64.85                                & 72.68                                & 55.31                                & 65.79                                & 61.88                                & 65.50                                 \\
      -                       & Deeplabv3+  \cite{deeplabv3p}              & 63.39                                & 69.97                                & 64.89                                & 68.65                                & 57.08                                & 59.46                                 \\
      -                       & UNet++    \cite{zhou2018unet++}                & 60.26                                & 66.88                                & 61.71                                & 71.68                                & 55.31                                & 58.43                                 \\ 
      \hline
      \multicolumn{8}{c}{Global-local scale patch based methods}                                                                                                                                                                                                                                                              \\ 
      \hline
      GLNet \cite{chen2019collaborative}                   & FPN                       & 51.71                                & 59.09                                & 58.69                                & 68.79                                & 21.86                                & 24.34                                 \\
      MagNet \cite{MagNet} & FPN &51.47&63.46&66.98&55.94&45.05 &49.09\\
      RAZN \cite{RAZN} &Deeplabv3+&58.62&66.27&63.52&68.49&57.63      &57.73\\                                                  
      WiCoNet \cite{WiCoNet}                 & ResNet50                  & 61.83                                & 75.68                                & 65.80                                & 76.72                                & 58.83                                & 70.28                                 \\
      CascadePSP \cite{cascadepsp}              & PSPNet                    & 67.48                                & 77.42                                & 72.45                                & 77.24                                & 67.35                                & 70.03                                 \\ 
      \hline
      \multicolumn{8}{c}{Scale-variable segmentation network GeoAgent}                                                                                                                                                                                                                                                                           \\ 
      \hline   
      GeoAgent & UNet           & 70.33$_{\textcolor{cyan}{(+16.23)}}$ & 77.11$_{\textcolor{cyan}{(+17.59)}}$ & 75.88$_{\textcolor{cyan}{(+17.12)}}$ & 75.23$_{\textcolor{cyan}{(+6.30)}}$  & 57.77$_{\textcolor{cyan}{(+4.75)}}$  & 61.37$_{\textcolor{cyan}{(+1.70)}}$   \\ 
      GeoAgent & UNet++           & 74.93$_{\textcolor{cyan}{(+14.67)}}$ & 79.17$_{\textcolor{cyan}{(+12.29)}}$ & 77.30$_{\textcolor{cyan}{(+15.59)}}$ & 78.62$_{\textcolor{cyan}{(+6.94)}}$  & 61.84$_{\textcolor{cyan}{(+6.53)}}$  & 60.01$_{\textcolor{cyan}{(+1.58)}}$   \\ 
      GeoAgent & Deeplabv3+           & 76.19$_{\textcolor{cyan}{(+12.80)}}$ & 80.56$_{\textcolor{cyan}{(+10.59)}}$ & 75.51$_{\textcolor{cyan}{(+10.62)}}$ & 76.07$_{\textcolor{cyan}{(+7.42)}}$  & 62.25$_{\textcolor{cyan}{(+5.17)}}$  & 63.21$_{\textcolor{cyan}{(+3.75)}}$   \\ 
      GeoAgent & PSPNet           & 75.19$_{\textcolor{cyan}{(+14.82)}}$ & 79.62$_{\textcolor{cyan}{(+10.33)}}$ & 75.57$_{\textcolor{cyan}{(+16.53)}}$ & 79.34$_{\textcolor{cyan}{(+10.00)}}$ & 69.67$_{\textcolor{cyan}{(+9.98)}}$  & 74.36$_{\textcolor{cyan}{(+10.09)}}$  \\ 
      GeoAgent & FPN          & 76.00$_{\textcolor{cyan}{(+11.15)}}$ & 82.46$_{\textcolor{cyan}{(+9.78)}}$  & 78.16$_{\textcolor{cyan}{(+22.85)}}$ & 77.56$_{\textcolor{cyan}{(+11.77)}}$ & 73.95$_{\textcolor{cyan}{(+12.07)}}$ & 74.11$_{\textcolor{cyan}{(+8.61)}}$   \\ 
      \hline
      \end{tabular}        
  }
\end{table*}

Three semantic segmentation datasets for HSR remote sensing images were used for experiments: 

\textbf{Gaofen Image Dataset (GID)} \cite{tong2020land}: GID is a high resolution semantic segmentation  dataset based on Gaofen-2 (GF-2) satellite images. 
The satellite collects spectral reflectance information in the red-green-blue and near-infrared ranges to create 4-band images, with a spatial resolution of 4 meters from the multispectral camera. The dataset includes 10 training and testing images, each labeled with 15 categories and with a size of $7200\times6800$. The GID dataset is split into 50\% for training and 50\% for testing.

\textbf{Five-Billion-Pixels (FBP)} \cite{tong2022enabling}: FBP is large-scale LULC semantic segmentation dataset. Similar to the GID dataset, the FBP dataset is built based on 4-m resolution GF-2 images annotated in a 24-category system.
The FBP dataset contains 150 images with 5 billion labeled pixels, and the size of each image is also $7200\times6800$.

\textbf{Wuhan urban semantic understanding dataset (WUSU)}: We create A high-resolution LULC semantic segmentation dataset for Wuhan using GF-2 satellite imagery with finer annotations. The spatial resolution of the WUSU dataset was improved to 1-m through the fusion of panchromatic data. The categories were refined to differentiate between building instances and label them as high- or low-rise buildings, as well as to annotate excavations and structures within the city. The dataset annotates six satellite images with a total of 67,764 instances, with image heights and widthes ranging from 5500 pixels to 7025 pixels. The dataset can be found in https://github.com/AngieNikki/openWUSU.

\begin{figure*}[!htb] 
    \centering
    \includegraphics[width=0.9\linewidth]{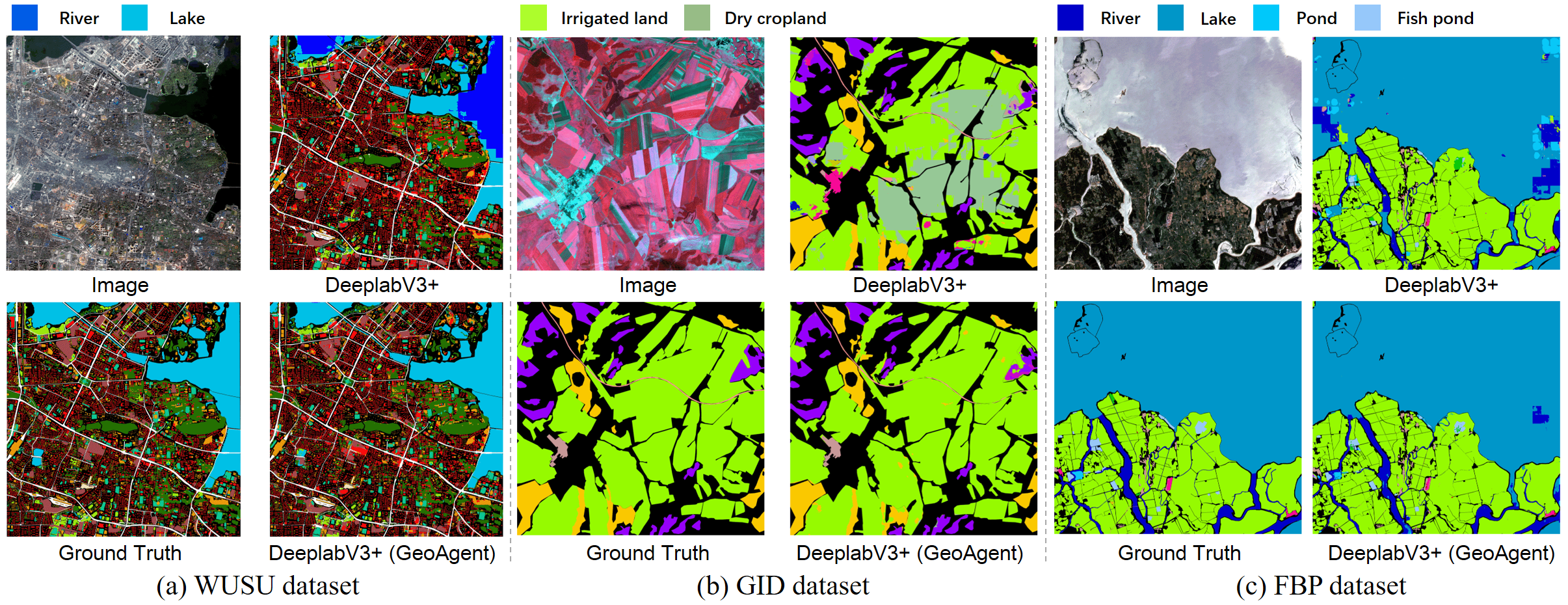}
    \vspace{-0.2cm}
    \caption{Segmentation results of Deeplabv3+ and modified GeoAgent models on three dataset.}
    \label{fig:Results}
    \vspace{-0.5cm}
\end{figure*}

The GeoAgent was implemented based on PyTorch and Stable-Baseline3 \cite{stable-baselines3}. 
The ResNet-18 is used for policy networks, and the backbones of segmentation networks are ResNet-50. 
The training patches were randomly sampled with the patch size of $512\times512$ and batch size 16. 
The stochastic gradient descent optimizer was used with the learning rate initialized to 0.001 with decay. The hyperparameters of A2C follow the default settings of stable-baseline3, except that the batch size is adjusted to 16. 
The models were trained for 20,000 steps on the GID and WUSU datasets and 60,000 on the FBP dataset. 
The segmentation \textit{agents} are pre-trained for 10,000 steps on the GID and WUSU datasets and 20000 on the FBP dataset.
In the joint training phase, the two components perform asynchronous parameter updates at 100-step intervals.
The mIoU and mean F$_1$ score were used for evaluation.
\vspace{-0.1cm}
\subsection{Comparison With Segmentation Methods}
\vspace{-0.1cm}
The proposed GeoAgent is a generic framework where the segmentation backbone can be an arbitrary FCN model. The GeoAgent is compared with the commonly used HSR remote sensing imagery semantic segmentation networks, including UNet \cite{unet}, UNet++ \cite{zhou2018unet++}, FPN \cite{FPN}, PSPNet \cite{pspnet}, Deeplabv3+ \cite{deeplabv3p}. Additionally, the global scale methods such as GLNet \cite{chen2019collaborative}, MagNet \cite{MagNet}, RAZN \cite{RAZN}, WiCoNet \cite{WiCoNet} and CascadePSP \cite{cascadepsp} are also compared.

The results listed in Table.\ref{table:results} show that GeoAgent significantly outperforms different patch-based based segmentation models in the challenging three HSR remote sensing image segmentation datasets.
With the introduction of the GeoAgent, a high accuracy improvement can be obtained for all the models. 
When compared to fixed global-local scale patch-based methods such as GLNet, WiCoNet, and CascadePSP, which also employ ResNet50, SFPN, and PSPNet as backbones, GeoAgent also demonstrates accuracy advantages on all three datasets.

Examples of Deeplabv3+ on three datasets are given in Fig.\ref{fig:Results}. For some artificial objects, such as roads, buildings, and residential areas, Deeplabv3+ can distinguish these categories well. However, larger natural objects, such as "rivers" and " in the WUSU and FBP datasets, have almost the same features as lakes and ponds under the finite size $512\times512$ image patch, and the "irrigated land" and "dry cropland" in the GID are also very confusing.
The fixed scale image patch limits all existing models to identify different parts of the water body as different classes, leading to fragmented segmentation results. 
In contrast, the proposed GeoAgent can obtain more global information by capture capturing context outside the image patch to a suitable size based on different objects, the large-scale LULC geo-objects are correctly identified and segmentation results are consistent.

\vspace{-0.1cm}
\subsection{Ablation Experiment}
\vspace{-0.1cm}
In this section, the ablation study was conducted to verify the validation of the proposed GeoAgent. Different scale-selection policies of SCA are compared. All experiments are based on the Deeplabv3+, benefiting from its balanced performance and computational complexity. The GeoAgent was compared with the following strategies:

\textbf{Local only}: This baseline policy uses the traditional pipeline of FCNs with fixed $512\times512$ local patches as input.
\textbf{Context only}: In this policy, the datasets were down-sampled with fixed scales by 2-6 times, and the $512\times512$ down-sampled context patches were directly fed into FCNs.
\textbf{Single-branch}: Although the SCA module is used to select the appropriate patch scale, only one branch, a normal segmentation network, is used to process the different multi-scale image patches. This policy is similar to RAZN \cite{RAZN} 
\textbf{Fixed scale}: The local and context branches were trained with the local image patches and context thumbnails, but the scales of the context branch are fixed to 2-6. This policy is similar to the fixed scale methods GLNet \cite{chen2019collaborative} and WiCoNet\cite{WiCoNet}.
\textbf{Random scale}: 
In this policy, the local and context branches were both activated, but the scales were randomly selected. Actions were randomly performed five times to report the mean and variance. This is to verify that the DRL-based scale selection gives better results than the random scale selection. When testing, the model performs inference at all scales, and the obtained probabilities are averaged to obtain the final results.

\begin{figure}[!htb] 
    \centering
    \begin{subfigure}{0.47\linewidth}
        \includegraphics[width=\linewidth]{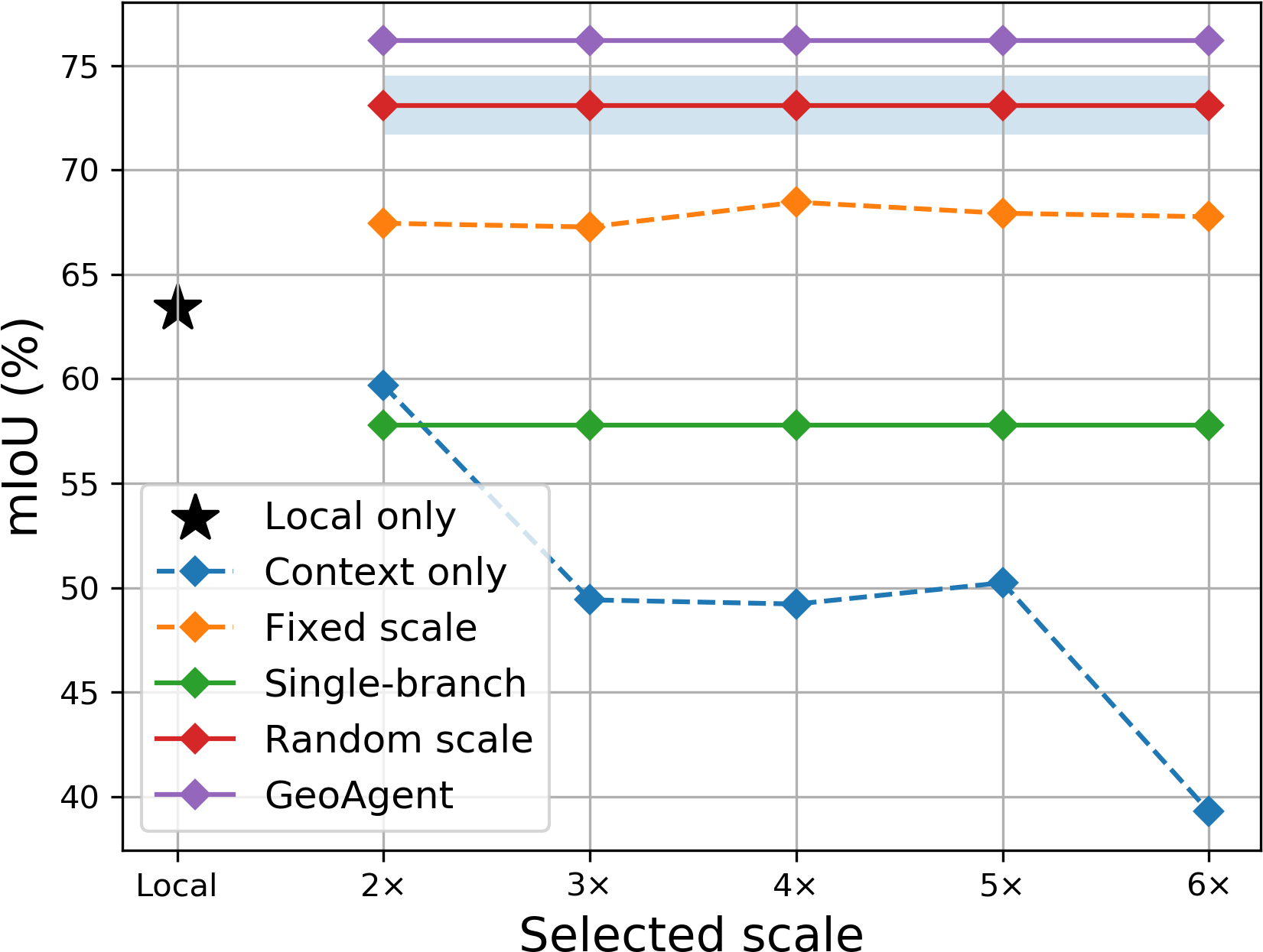}
        \caption{WUSU dataset}
        \label{fig_Scales_WUSU}
    \end{subfigure}
    \hfill
    \begin{subfigure}{0.49\linewidth}
        \includegraphics[width=\linewidth]{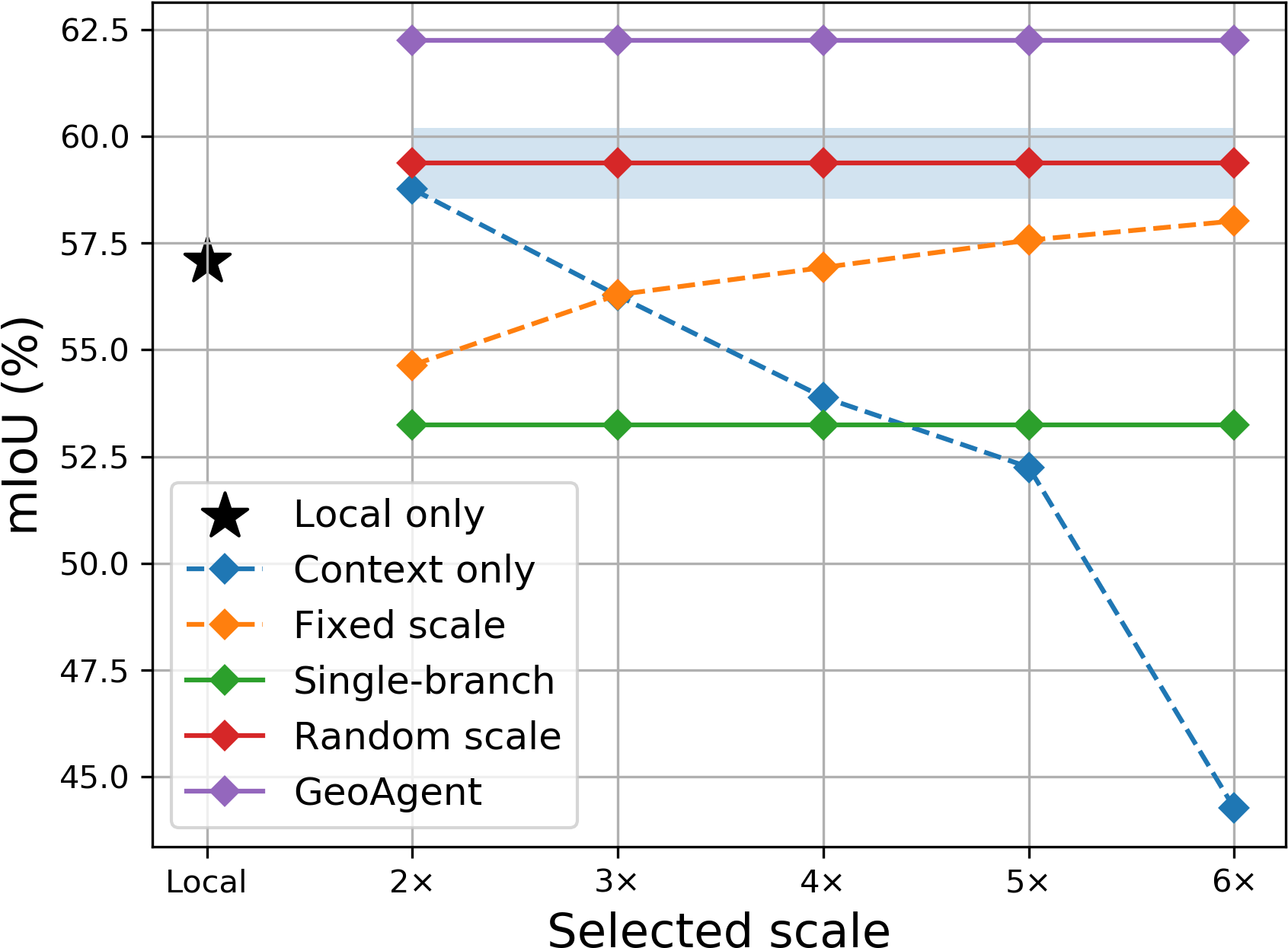}
        \caption{FBP dataset}
        \label{fig_Scales_FBP}
    \end{subfigure}    
    \vspace{-0.3cm}
    \caption{Ablation experiment results on different datasets.}
    \label{fig:Scales}
    \vspace{-0.3cm}
  \end{figure}

The results for the WUSU and FBP datasets are shown in Fig.\ref{fig:Scales}. The segmentation accuracy of the simple downsampled thumbnails decreases significantly as the selected scale increases, which is reasonable because the increasingly large scale downsampling directly destroys the information in the images. The global information introduced using the fixed scale performs better but negatively affects the final segmentation results in specific cases (fixed 2x scale for the FBP dataset). Moreover, the optimal fixed scales are different for different datasets. Random scale policy performs better, with a significantly higher final score than the baseline using only local information. Our proposed GeoAgent, on the other hand, dynamically decides the scale of patches and achieves the best performance.

\vspace{-0.1cm}
\subsection{Sensitivity Analysis of SCA}
\begin{figure}[!h]
    \centering
    \begin{subfigure}{0.49\linewidth}
      \includegraphics[width=\linewidth]{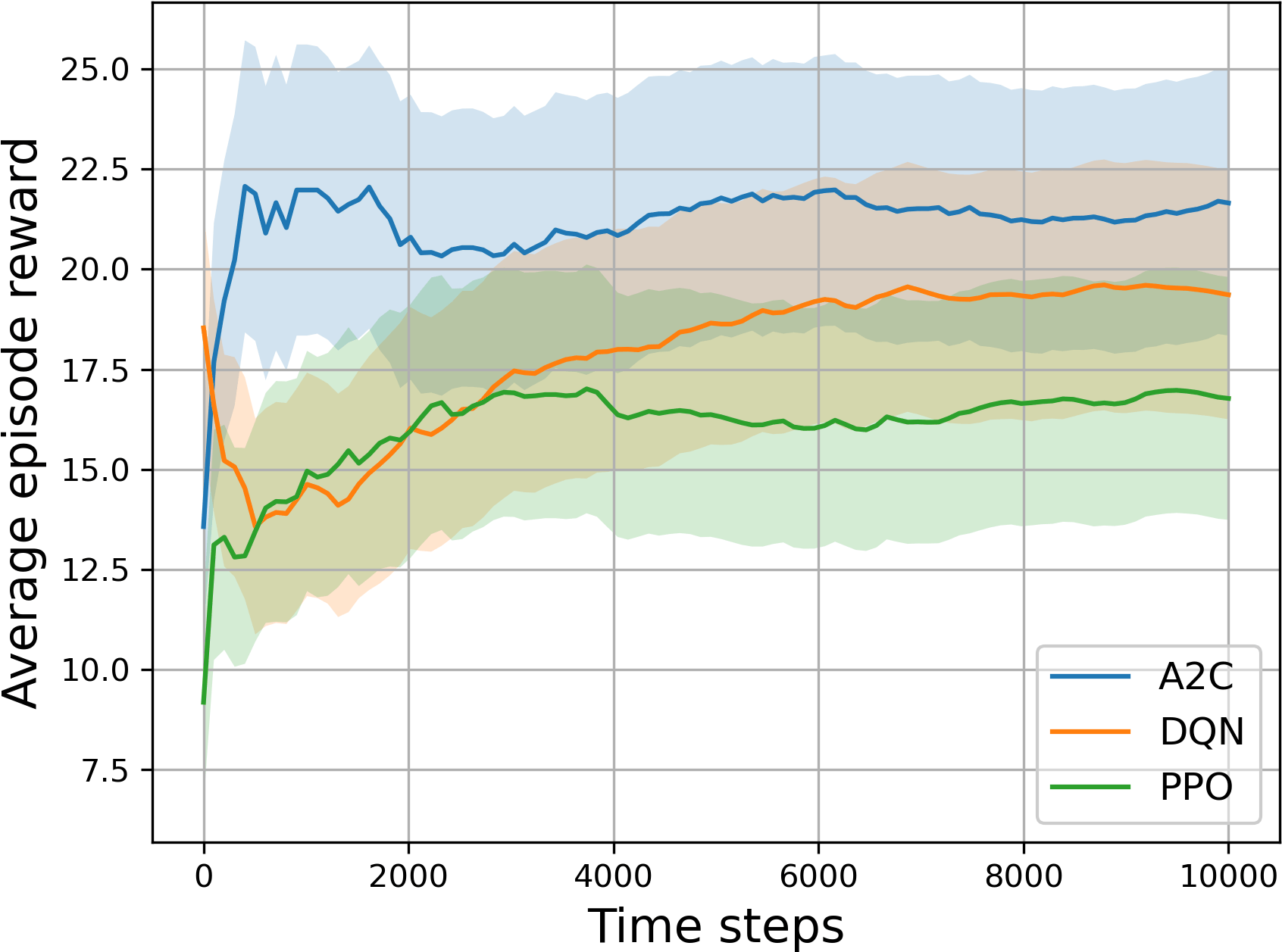}
      \caption{WUSU dataset}
      \label{fig_DRLs_WUSU}
    \end{subfigure}
    \hfill
    \begin{subfigure}{0.49\linewidth}
        \includegraphics[width=\linewidth]{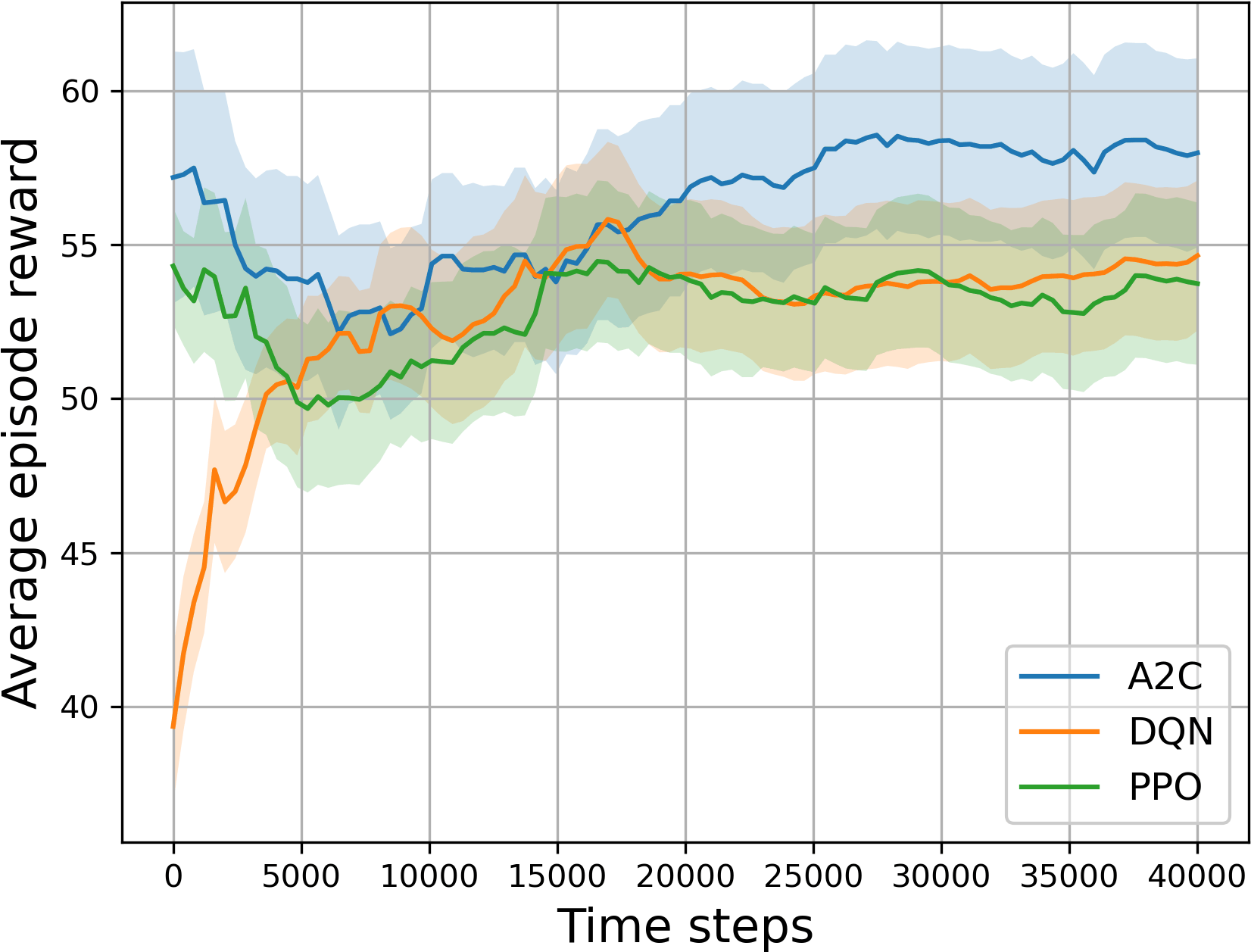}
        \caption{FBP dataset}
        \label{fig_DRLs_FBP}
    \end{subfigure}    
    \vspace{-0.3cm}
    \caption{Average performance of various DRL methods in different datasets.}
    \label{fig:DRLs}
    \vspace{-0.3cm}
\end{figure}

\vspace{-0.1cm}
In this section, we investigate the impact of various reinforcement learning methods and backbones on the policy network of GeoAgent. As the FCN and DRL algorithms of the GeoAgent framework are interchangeable, we train all methods compared in Section 4.2 using different DRL algorithms for each dataset.  In addition to the A2C used \cite{A2C}, two representative algorithms, Deep Q Network (DQN) \cite{mnih2015human} and Proximal Policy Optimization (PPO) \cite{PPO}, are employed for comparison. The means and standard deviations of average episode \textit{rewards} during training with different DRL algorithms for each dataset are shown in Fig.\ref{fig:DRLs}. 

All DRL algorithms prove effective in learning scale-selection policies on all three datasets. On the all datasets, all models converge rapidly due to the relatively small size of the data volume, with A2C converging the fastest and ultimately converging to the highest value of the average episode \textit{rewards}.  A2C demonstrates an advantage in terms of convergence speed and results, although other DRL methods also perform well.
\par
The models may prioritize short-term returns over the long-term mapping accuracy of the entire image, which is represented by $\mathcal{R}_{map}$. To address this, the accuracy gain of different models using GeoAgent is calculated and presented in Table.\ref{table:DRLs}, in addition to the average episode returns during training. GeoAgent consistently delivers significant mapping accuracy gains on all datasets. While there are some differences in performance among the DRL methods, they all result in significant accuracy gains. 
Despite these differences, the final mapping results are not significantly impacted, highlighting the robustness of the GeoAgent framework to reinforcement learning algorithms.

\begin{table}[htb]
    \centering
    \small
    \tabcolsep=0.2cm
    \vspace{-0.0cm}
    \begin{tabular}{cccc}
     \hline
     Methods       & WUSU & GID & FBP \\ 
     \hline
     DQN \cite{mnih2015human}         &   13.85$\pm$1.9 &15.63$\pm$3.0 &\textbf{7.76$\pm$3.0}   \\
     PPO \cite{PPO}        &   13.51$\pm$1.8 &15.77$\pm$3.0 &7.76$\pm$3.0   \\
     A2C \cite{A2C}       &   \textbf{13.93$\pm$1.7} &\textbf{16.54$\pm$3.1} &7.70$\pm$2.9    \\
     \hline
     \end{tabular}
    \vspace{-0.1cm}
     \caption{Accuracy improvement of various FCNs for different DRL methods in the three datasets. 
     }
     \vspace{-0.3cm}
     \label{table:DRLs}
\end{table}

\vspace{-0.1cm}
\subsection{Visualization Analysis}
\begin{figure}[!htb]
  \centering
  \includegraphics[width=\linewidth]{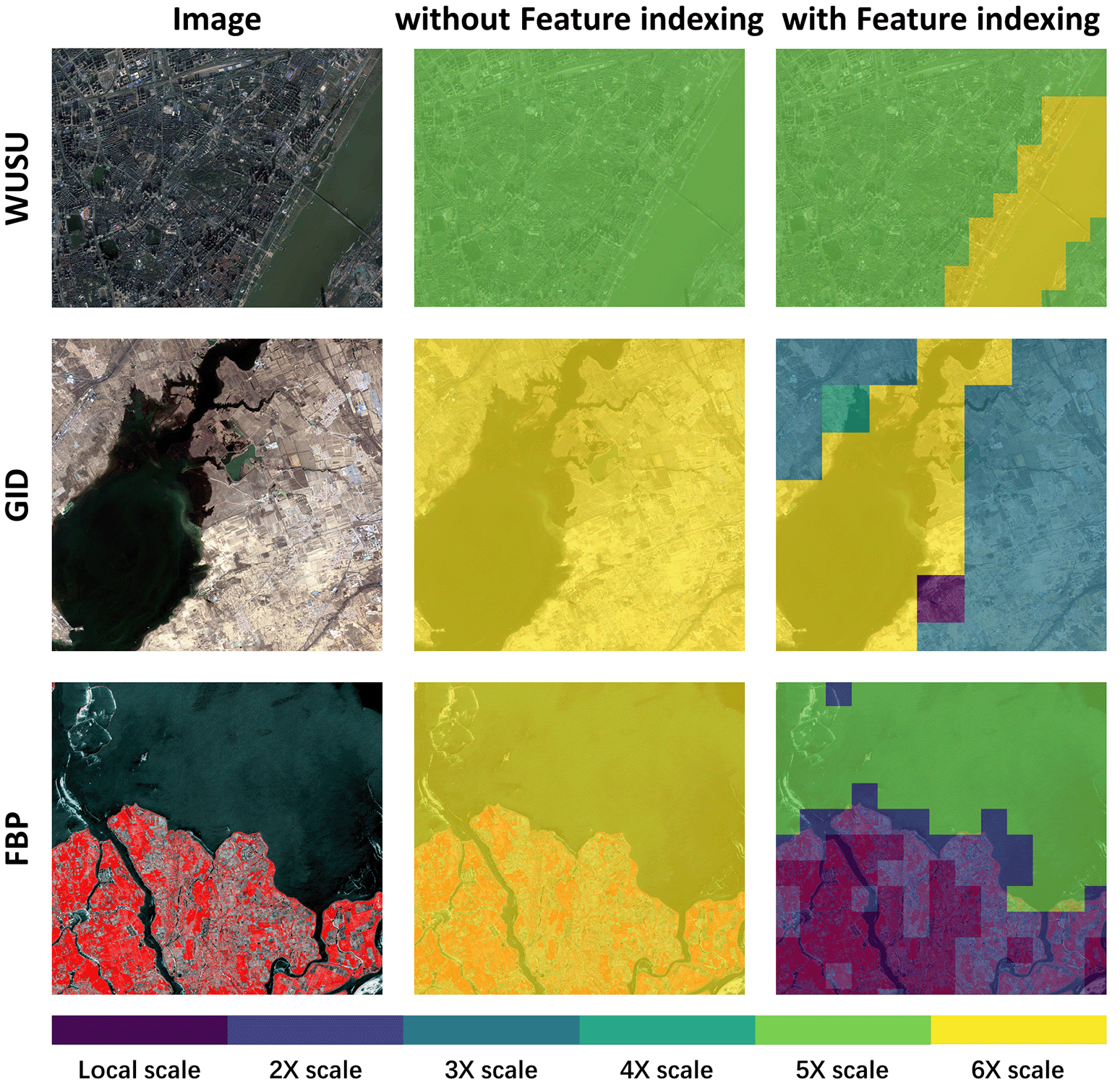}
  \vspace{-0.5cm}
  \caption{Action map visualization of GeoAgent on different datasets.}
  \label{fig:Visualization}
  \vspace{-0.3cm}
\end{figure}

\vspace{-0.1cm}
This section analyzes the \textit{actions} of SCA, which are recorded by sliding a window through an image and completing an episode. The resulting \textit{action} maps of GeoAgent models based on A2C and Deeplabv3+ are presented in Fig. \ref{fig:Visualization} for three datasets. To evaluate the effectiveness of the proposed feature indexing method, the \textit{action} maps of the policy network without this module are compared. In this case, thumbnails and position mask \textit{states} are concatenated to form a multi-channel image, which is then fed to the CNN for feature extraction and output \textit{action} probabilities.
\par
\par
The results indicate that the GeoAgent can adaptively take different context scale for different \textit{states}. For large-scale objects, such as rivers and lakes, larger scale chosen to obtain more background information due to the lack of long-range spatial relationships at smaller observation scales. The SCA mainly take larger magnification zoom for natural objects with large scales, while smaller scales or even no zooming are chosen for artificial features that can be easily identified.

Without the feature indexing module, the policy network finds it difficult to locate the position of the current \textit{state} on the whole thumbnail, as the position mask is fed in only as a channel of observation, and the thumbnails of different \textit{states} are identical. Consequently, the policy network without feature indexing takes the same \textit{action} for all \textit{states} on the whole image. In the two examples of large-scale objects on the GID dataset and the FBP dataset, the conservative \textit{action} of taking the largest scale on the whole image is chosen by the \textit{agent} to guarantee a higher average expected return. This outcome highlights the effectiveness of the proposed feature indexing method.
\par
\begin{figure}[htb]
   \centering
    \includegraphics[width=1.0\linewidth]{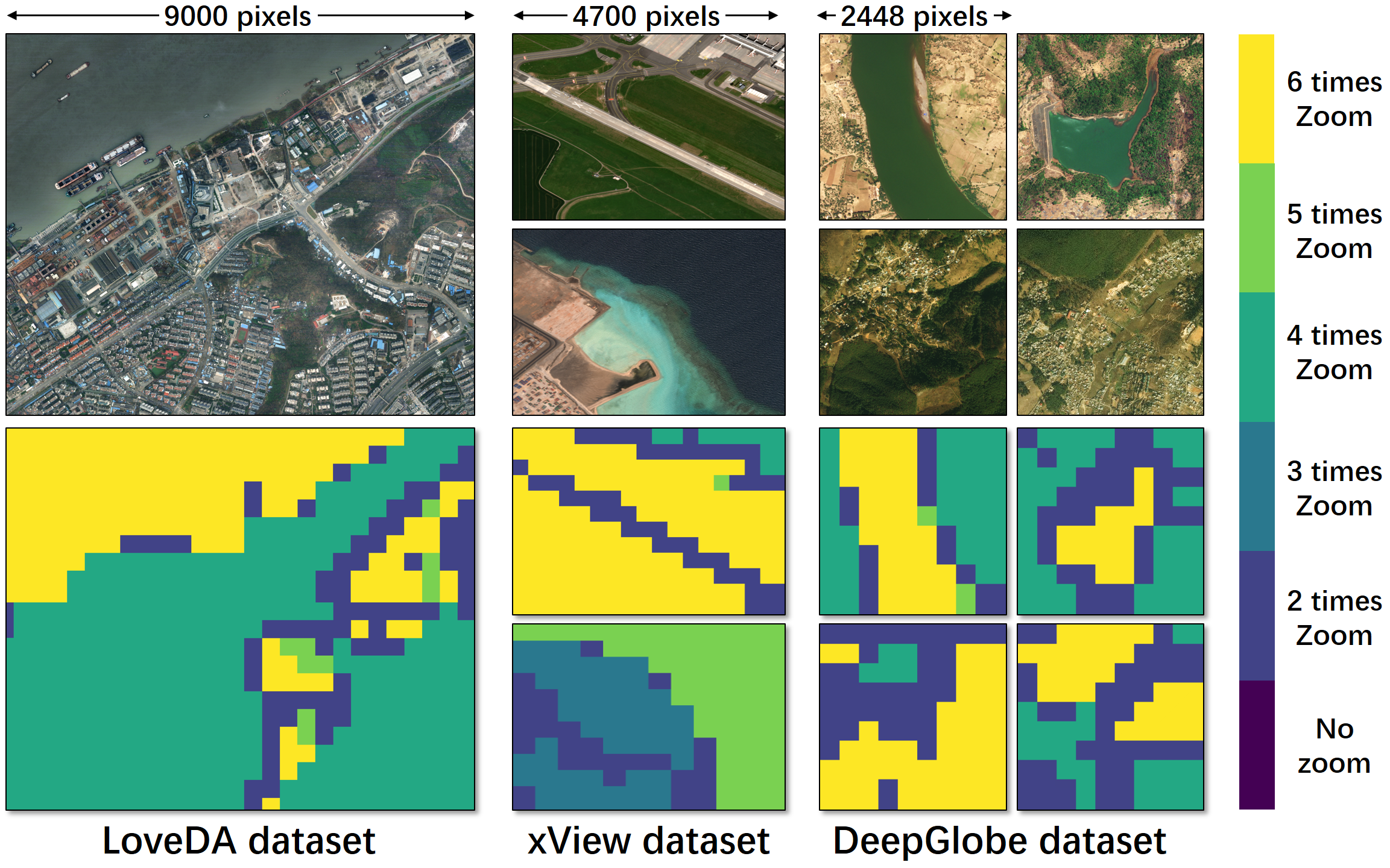}
    \vspace{-0.7cm}
    \caption{Action visualization of the GeoAgent on unseen datasets.}
    \vspace{-0.5cm}
    \label{fig:other}
\end{figure}

In addition, to evaluate the generalization of GeoAgent, we trained an Agent with RGB input on the WUSU dataset and performed inference on unseen WorldView-based datasets, namely LovaDA \cite{wang2021loveda}, xView\cite{lam2018xview}, and DeepGlobe\cite{demir2018deepglobe}. The action maps in Fig.\ref{fig:other} demonstrate GeoAgent's ability to learn action strategies that generalize well across different datasets. \\
GeoAgent consistently chooses larger scales for big objects to understand their context, while it selects localized scales for smaller objects, showing its ability to focus on detail. This pattern is observed regardless of the dataset's origin, proving its effective generalization skills.


\vspace{-0.4cm}
\vspace{-0.1cm}
\section{Conclusion}
\vspace{-0.1cm}
In this paper, we propose a deep reinforcement learning-based scale-adaptive semantic segmentation network called GeoAgent, breaking the limitation sliding window size of that conventional segmentation networks. Experiments on three datasets show that our proposed GeoAgent achieves state-of-the-art results on the semantic segmentation task of high-resolution satellite images.
\section{Acknowledgements}
This work was supported by National Natural Science Foundation of China under Grant No. 42325105, 42071350, and LIESMARS Special Research Funding.

{\small
\bibliographystyle{ieee_fullname}
\bibliography{DRLtrans}
}

\end{document}